\documentclass[conference,10pt]{IEEEtran}
\usepackage[utf8]{inputenc}
\def\BibTeX{{\rm B\kern-.05em{\sc i\kern-.025em b}\kern-.08em
    T\kern-.1667em\lower.7ex\hbox{E}\kern-.125emX}}

\usepackage{graphicx}
\usepackage{color}
\usepackage{cite}
\usepackage{amsmath}
\usepackage[caption=false]{subfig}
\usepackage{amssymb}
\usepackage{comment}
\usepackage{mathtools}
\usepackage{tabulary}
\usepackage{acronym}
\usepackage{upgreek}
\usepackage{algorithm}
\usepackage{algpseudocode}
\usepackage{placeins}
\usepackage[super]{nth}
\usepackage{dsfont}
\usepackage{hyperref}
\usepackage{flushend}

\newcommand{\ul}[1]{\underline{#1}}
\newlength{\figwidth}
\title{Graph-CNNs for RF Imaging:\\ 
Learning the Electric Field Integral Equations
}

\author{
    \IEEEauthorblockN{
        Kyriakos Stylianopoulos$^1$, Panagiotis Gavriilidis$^1$, 
        Gabriele Gradoni$^2$, and George C. Alexandropoulos$^1$
    }
    \IEEEauthorblockA{
        $^1$Department of Informatics and Telecommunications,
        National and Kapodistrian University of Athens,\\
        Panepistimiopolis Ilissia, 16122 Athens, Greece
    }
    \IEEEauthorblockA{
        $^2$Institute for Communication Systems, University of Surrey, Guildford GU27XH, UK\\
        e-mails: \{kstylianop, pangavr, alexandg\}@di.uoa.gr, g.gradoni@surrey.ac.uk
    }
}

\begin{document}

\maketitle

\begin{abstract}
Radio-Frequency (RF) imaging concerns the digital recreation of the surfaces of scene objects based on the scattered field at distributed receivers. To solve this difficult inverse scattering problems, data-driven methods are often employed that extract patterns from similar training examples, while offering minimal latency.
In this paper, we first provide an approximate yet fast electromagnetic model, which is based on the electric field integral equations, for data generation, and subsequently propose a Deep Neural Network (DNN) architecture to learn the corresponding inverse model.
A graph-attention backbone allows for the system geometry to be passed to the DNN, where residual convolutional layers extract features about the objects, while a UNet head performs the final image reconstruction.
Our quantitative and qualitative evaluations on two synthetic data sets of different characteristics showcase the performance gains of thee proposed advanced architecture and its relative resilience to signal noise levels and various reception configurations. 
\end{abstract}

\let\thefootnote\relax\footnotetext{This work has been supported by the SNS JU project 6G-DISAC under the EU’s Horizon Europe research and innovation program under Grant Agreement No 101139130 and the 2023 Greek-British Cooperation on Short-Term
Scholarships Programme.}


\section{Introduction}

Radio-Frequency (RF) imaging is widely used in many areas, such as in medical diagnostics~\cite{Microwave_Breast_Imaging}, security applications~\cite{RMA}, and diverse engineering fields, such as structural tomography~\cite{MAIERHOFER2010xv} and ground-penetrating radar~\cite{GPR_book}.
Multi-static imaging setups, where multiple Transmitters (TXs) emit RF waves that interact with surrounding objects before being recorded by distributed Receivers (RXs), enable imaging through digital signal processing or techniques realized directly in the Electromagnetics (EM) domain. However, typical systems require specialized hardware, high computational resources, and operate in high power and bandwidth regimes.

Next generation wireless networks are expected to offer imaging opportunities through communications infrastructure, leveraging base stations densification, device-to-device networking, and millimeter-wave technology. 
Applications such as integrated sensing and communications~\cite{6G-DISAC-magazine}, digital twinning~\cite{MVA24_Metaverse}, and autonomous guided vehicles benefit from high-resolution channel propagation characterization~\cite{DT_for_Wireless} and optimized beamforming~\cite{QUALCOMM_channel_model}.
Integrating, however, imaging functionalities into wireless networks imposes stringent requirements on latency, computation, and communication resources, challenging existing methodologies.

In RF imaging, Synthetic Aperture Radar (SAR) techniques, such as the range migration algorithm~\cite{RMA, BPA_vs_RMA, RMA_MIMO}, reconstruct scenes by coherently combining multiple received signals across frequency or spatial domains.
While enabling high-resolution imaging, SAR requires large bandwidth, struggles in near-field scenarios, and relies on simplified reflection models that overlook diffraction and multi-bounce effects.
Moreover, SAR primarily captures outer surfaces, making it ineffective for occluded objects within a Domain of Interest (DoI).
Alternatively, RF imaging can directly leverage EM theory, capitalizing on the fact that received signals satisfy Maxwell’s frequency-domain equation~\cite{Yonina_Physics_ML_Imaging}, inherently encapsulating the DoI’s EM properties.
This approach circumvents SAR's limitations, but presents its own challenges: solving nonlinear Partial Differential Equations (PDEs) is computationally expensive, prone to degenerate solutions due to ill-posedness, and depends on numerical models lacking accuracy.

Machine Learning (ML) methodologies provide an attractive alternative in overcoming such shortcomings.
The plethora of real life data sets of medical data has been instrumental to the successful applications of Deep Neural Networks (DNNs) in medical imaging (see~\cite{Yonina_Deep_Tomographic_Imaging} and references therein) through the exploitation of diverse computer vision schemes, such as image denoising, super resolution, and unsupervised learning.
Additionally, ample synthetic data sets may be generated in simulations using EM models, which provide a stream of training examples with de facto labeling, to train ML approaches for general-purpose inverse scattering problems~\cite{Yonina_Physics_ML_Imaging}.
As a prominent example, the work in~\cite{UNet_for_Imaging} deployed the UNet architecture~\cite{U_Net} to capitalize on the DNN's ability to output images comprised of distinct regions. 
More recently, there has been an ongoing interest in combining ML with EM models to the purpose of designing Physics-Informed Neural Networks (PINNs), which enhance the fidelity of image reconstruction.
This may be achieved through pre- or post-processing of the input and output variables of DNN with physics-compliant models, by incorporating loss functions that conserve EM properties (e.g.,~\cite{DL_Imaging_Two_Step}), or by approximating parts of EM models with DNN layers; refer to~\cite{Yonina_Physics_ML_Imaging} for an in-depth analysis of PINNs for RF imaging. However, despite the promising results of PINNs in performance gains and data efficiency, there remain yet unresolved challenges.
First, the physics-compliant loss functions, or operations, are typically computationally expensive to evaluate, especially if such computations involve evaluating or solving PDEs.
Additionally, while one is free to choose from a wide variety of approximate EM models and their corresponding properties to infuse into PINNs, it is difficult to determine a priori which of those will be beneficial for training and what DNN layers are effective in such regimes.

In this paper, we argue that developing efficient DNN models with pertinent training methodologies, that better exploit all the available system information, is of paramount importance, either as standalone physics-agnostic imaging algorithms, or as components of PINNs.
The specific contributions of this work are as follows:
An approximate yet computationally fast EM model is first presented, which is based on the Electric Field Integral Equations (EFIEs) to produce the received signals scattered by objects within the DoI.
The model is then used to generate two synthetic RF imaging data sets addressing different challenges.
Next, a novel DNN architecture for RF imaging is proposed, which is based on graph neural networks that capture geometrical information about the system's transceivers and provide highly expressive feature extraction, followed by ResNet~\cite{ResNet} and UNet structures for effective image generation.
This architecture is evaluated on the generated data sets under varying Signal-to-Noise Ratio (SNR) conditions and numbers of RXs to showcase its performance gains over simpler conventional DNN approaches.

\section{System and Problem Formulation}
\subsection{EM Modeling}\label{sec:em_model}

We consider a 2D square DoI of radius $\rho = 5\lambda$, with $\lambda$ being the wavelength of the propagating electric field, which is centered at the origin.
$N_r$ RX units are placed in equally spaced positions around the periphery of the DoI, which are assumed to be perfectly synchronized.
Contrary to most RF imaging techniques~\cite{Yonina_Physics_ML_Imaging, DL_for_Inverse_Scattering_review, UNet_for_Imaging, DL_Imaging_Two_Step}, we assume a single TX located at a distance of $20\lambda$ from the origin, at \(\mathbf{p}_t=[20\lambda,0]^\top\).
Despite making the problem more challenging, using a single TX greatly simplifies the involved synchronization processes and reduces the measurement latency, offering a more attractive scenario for wireless communication systems.

Inside the DoI, there exist one ore more disjoint objects of perfect electrical conductors, that are represented as a discrete set of infinitesimal elements of coordinates $\boldsymbol{\bar{s}}=[x_{\bar{s}},y_{\bar{s}}]^\top$, sampled with spatial resolution $\rho_0$ inside the boundaries of each object.
Let $\mathcal{S}$ be the set of all potential points in the DoI, and $\mathcal{\bar{S}}$ be the set of all coordinates comprising the objects, which will be henceforth referred to as a surface, despite containing possibly disjoint objects.
We assume the transmitting field to travel in the \((x,y)\)-plane, being polarized in the \(z\)-direction, as
$E_{\rm t}(\boldsymbol{p})=E_0 {\exp({-\jmath \mathbf{k}^\top (\boldsymbol{p}-\boldsymbol{p}_{\rm t}})})/({4 \pi \|\boldsymbol{p}-\boldsymbol{p}_{\rm t}\|})$
with $\boldsymbol{p}=[x_p,y_p]^\top$ being an arbitrary point in the DoI, \(\mathbf{k} \in \mathbb{R}^{2\times 1}\) denoting the free space wave vector, and $E_0$ being a constant amplitude term. 
Following first physics principles and considering time-harmonic electric field (i.e., \(\mathcal{E}(t,\boldsymbol{p})=E_0\cos(\omega t-\mathbf{k}^\top(\boldsymbol{p}-\boldsymbol{p}_{\rm t}))\)), we formulate the relationship that relates the electric field at any point in space with the current density \(J(\boldsymbol{s})\) of the DoI via the EFIE, as follows:
\begin{equation}\label{eq:EFIEs}
    E_{\rm r}(\boldsymbol{p})=\iint\limits_{\mathcal{S}} G(\boldsymbol{p}, \boldsymbol{s}) J(\boldsymbol{s}) d\boldsymbol{s} + E_{\rm t}(\boldsymbol{p}),
\end{equation}
where $G(\boldsymbol{p}, \boldsymbol{s})$ is the Green's function.
For the $\text{TM}_z$ propagating wave considered here, this function is defined via the Hankel function of the $0$-th order of the $2$nd kind ($H_0^{(2)}$) as $G(\boldsymbol{p}, \boldsymbol{s}) = -\frac{\jmath}{4}H_0^{(2)}(k_0||\boldsymbol{p} - \boldsymbol{s}||)$\cite[eq.~(11-11a)]{Balanis}, where $k_0$ denotes the free-space wavenumber.
Computing the electric field reduces to determining the current density, which can be done by adopting $N$ (known) orthogonal basis functions $\{\Phi_n(\boldsymbol{s})\}_{n=1}^N$ and accompanying intensities $\{\alpha_n\}_{n=1}^N$. The current density is then expressed as 
$J(\boldsymbol{s}) = \sum_{n=1}^N \alpha_n \Phi_n(\boldsymbol{s})$.

To derive a complete EM model involves finding appropriate values of $\{\alpha_n\}_{n=1}^N$ for a given $\mathcal{\bar{S}}$, which can be obtained via the point-matching method \cite[Section~12.2.4]{Balanis}.
It is first noted that~\eqref{eq:EFIEs} can be applied to any point in 2D space, including the RXs, empty space, as well as $\boldsymbol{\bar{s}} \in \mathcal{\bar{S}}$ in particular.
We thus exploit the fact that the set $\mathcal{\bar{S}}$ contains perfect electric conductors, which dictates the following boundary condition onto the surface points:
\begin{equation}\label{eq:boundary-condition}
    E_{\rm r}(\boldsymbol{\bar{s}}) = 0, ~\forall \boldsymbol{\bar{s}} \in \mathcal{\bar{S}}.
\end{equation}
By assuming known $\Phi_n(\boldsymbol{s})$ and by evaluating~\eqref{eq:EFIEs} on $N$ different $\boldsymbol{\bar{s}}_i$, $\forall i=1,\ldots,N$ points, we obtain a linear system of $N$ equations with $N$ unknowns $\mathbf{a} \triangleq [\alpha_1, \dots, \alpha_N]^\top$ through~\eqref{eq:boundary-condition}.
Specifically, let 
$\mathbf{e}_{\rm t} \triangleq [-E_{\rm t}(\boldsymbol{\bar{s}}_1),\dots,-E_{\rm t}(\boldsymbol{\bar{s}}_N)]^\top$ and the $N\times N$ matrix $\mathbf{M}$ as
$[\mathbf{M}]_{i,n} \triangleq \iint_{\mathcal{S}} G(\boldsymbol{\bar{s}}_i, \boldsymbol{s}) \Phi_n(\boldsymbol{s}) d\boldsymbol{s}$ $\forall i,n = 1,\dots,N$.
Assuming $\mathbf{M}^{-1}$ is defined, the system of equations deduced from~\eqref{eq:boundary-condition} can be solved as $\mathbf{a} = \mathbf{M}^{-1} \mathbf{e}_{\rm t}$, thus, completing the model.
For the data-generation process we have selected the following procedure for the creation of $\Phi_n$:
An $\sqrt{N} \times\sqrt{N}$ grid of equally spaced points $\boldsymbol{\hat{s}}_n$ is constructed.
Each basis $\Phi_n$ is defined as a 2D square pulse of unit amplitude centered around $\boldsymbol{\hat{s}}_n$ with a width of $2\rho/\sqrt{N}$.

\subsection{RF Imaging Problem Formulation}
For each instance \(t\) of the problem, the DoI contains a distinct surface $\mathcal{\bar{S}}^{(t)}$.
Let the (fixed) positions of the RXs be denoted as $\boldsymbol{r}_k = [x_{r_k}, y_{r_k}]^\top$ $\forall k=1,\ldots,N_r$, and the corresponding vector of received signals as $\boldsymbol{e}^{(t)} \triangleq [E_{\rm r}(\boldsymbol{r}_1), \dots, E_{\rm r}(\boldsymbol{r}_{N_r})]^\top$.
The model described in Section~\ref{sec:em_model} is referred to as the {\em forward} model, as it provides the mapping $F:\mathcal{\bar{S}}^{(t)}  \to \boldsymbol{e}^{(t)}$.
As a general formulation, the imaging objective can be described as the optimization problem
$\mathcal{S}_{\rm opt} \triangleq {\rm arg}\max_{\mathcal{\bar{S}}}\|\boldsymbol{e}^{(t)} - F(\mathcal{\bar{S}})\|$, which seeks the surface that, when processed through the {\em forward} model, best matches the received electric fields.
However, the above is analytically intractable 
involving searching over the space of all possible surfaces which grows exponentially with DoI's resolution.

To this end, a data-driven objective is employed instead.
For notation purposes, let us first define the operator $I(\mathcal{\bar{S}})$ that transforms the surface $\mathcal{\bar{S}}$ to a discretized binary image of the DoI of arbitrary spatial resolution, where each pixel has the value of $1$ if and only if there is at least one point in $\mathcal{\bar{S}}$ that is spatially covered by the pixel.
Assume that $\mathcal{\bar{S}}^{(t)}$'s elements are conditionally dependent and are thus sampled by a non-uniform distribution, which motivates the need for ML applications. 
The learning objective can be defined as approximating the {\em inverse} model $F^{-1}: \boldsymbol{e}^{(t)} \to \mathcal{\bar{S}}^{(t)}$ as:
\begin{align}\label{eq:inverse-problem}
    F^{-1}_{\rm opt} &\triangleq {\rm arg}\min_{F^{-1}} \frac{1}{|\mathcal{D}|}\sum_{t=1}^{|\mathcal{D}|}{\rm CE}(I(\mathcal{\bar{S}}^{(t)}), I(F^{-1}(\boldsymbol{e}^{(t)})),
\end{align}
where the Cross-Entropy (CE) function ${\rm CE}(\boldsymbol{x}, \boldsymbol{y}) = - \sum_{j} [\boldsymbol{x}]_j \log [\boldsymbol{y}]_j$  is the objective function ($I(\cdot)$ is treated as a vector) and $\mathcal{D}$ is a collected set of input-target data tuples $\{ \boldsymbol{e}^{(t)}, I(\mathcal{\bar{S}}^{(t)})\}$.
For ML approaches, $F^{-1}(\cdot)$ is a parameterized neural network, hence, problem~\eqref{eq:inverse-problem} can be solved through Stochastic Gradient Descent (SGD), by optimizing the DNN's trainable weight matrices over batches of set $\mathcal{D}$.


\section{Proposed Deep Neural Network Architecture}\label{sec:dnn-arch}

\begin{figure}
    \centering
    \includegraphics[width=\linewidth]{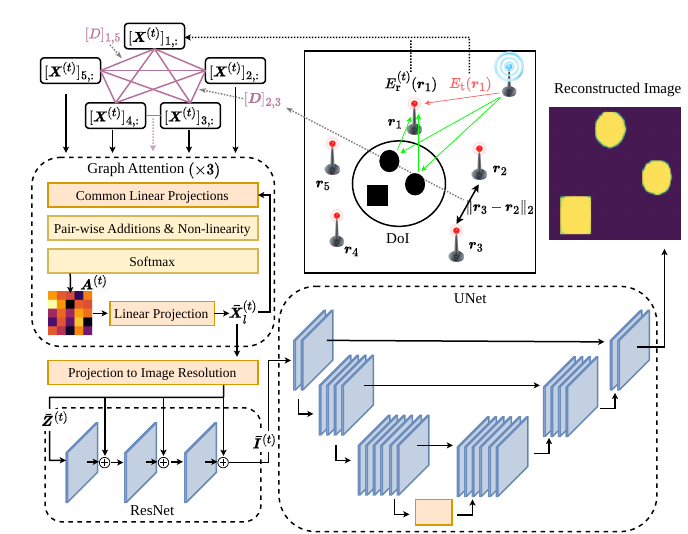}
    \caption{GAT-Res-UNet architecture for RF imaging via inverse scattering.}
    \label{fig:dnn-arch}
\end{figure}

The proposed DNN architecture consists of three parts.
First, a Graph-Attention Network (GAT)~\cite{GAT, pyg_gatconv} is used for feature extraction.
The imposed graph structure allows for the incorporation of systemic and geometric information (that remains fixed throughout the data set) as node and edge features, while the attention mechanism is tasked to learn correlations between the electric field of each data instance inputs and those system variables.
In particular, let the input data matrix for the $k$-th data instance $\boldsymbol{X}^{(t)} \in \mathbb{R}^{N_r \times 4}$ contain the received field at the RXs ($\boldsymbol{e}^{(t)}$) and the (fixed) electric field contribution solely by the TX as
$[\boldsymbol{X}^{(t)}]_{i,:} \triangleq [\mathfrak{Re}(E^{(t)}_{\rm r}(\boldsymbol{r}_i))~ \mathfrak{Im}(E^{(t)}_{\rm r}(\boldsymbol{r}_i))~ \mathfrak{Re}(E_{\rm t}(\boldsymbol{r}_i))~ \mathfrak{Im}(E_{\rm t}(\boldsymbol{r}_i))]^\top$.
Define the edge weights between the $i$-th and $j$-th RXs via their Euclidean distance as $[\boldsymbol{D}]_{i,j}\triangleq\| \boldsymbol{r}_i - \boldsymbol{r}_j\|_2$.
The attention coefficients of the first GAT layer $[\boldsymbol{A}^{(t)}]_{i,j}$ are computed as:
\begin{align}
    [\boldsymbol{A}^{(t)}]_{i,j} \triangleq& {\rm softmax}\Big({\rm LeakyReLU}\big(\boldsymbol{w}_{\rm s}^\top \boldsymbol{\Theta}_{\rm s} [\boldsymbol{X}^{(t)}]^\top_{i,:} \nonumber \\
    &+ \boldsymbol{w}_{\rm t}^\top \boldsymbol{\Theta}_{\rm t} [\boldsymbol{X}^{(t)}]^\top_{j,:} + \boldsymbol{w}_{\rm e}^\top \boldsymbol{\Theta}_{\rm e} [\boldsymbol{D}]_{i,j} \big) \Big),
\end{align}
where $\boldsymbol{w}_{\rm s}, \boldsymbol{w}_{\rm t}, \boldsymbol{w}_{\rm e} \in \mathbb{R}^{d \times 1}$ and $\boldsymbol{\Theta}_{\rm s},\boldsymbol{\Theta}_{\rm t}, \boldsymbol{\Theta}_{\rm e} \in \mathbb{R}^{d \times 4}$ are trainable weights with an arbitrary embedding dimension $d$. 
The output of the GAT layer is given as
$[\boldsymbol{\bar{X}}^{(t)}]_{i,:} = \sum_{j=1}^{N_r} [\boldsymbol{A}]_{i,j} \boldsymbol{\Theta}_{\rm t} [\boldsymbol{X}^{(t)}]^\top_{i,:}$.
For the considered data sets, $L_1=3$ GAT layers with $d=32$ are applied in sequence so that their respective inputs are $\boldsymbol{X}_{l}^{(t)} = \boldsymbol{\bar{X}}_{l-1}^{(t)}$ $\forall l=2,\dots,L_1$.

In the sequel, $\boldsymbol{X}_{L_1}^{(t)}$ is flattened to $\boldsymbol{z}^{(t)} \in \mathbb{R}^{dN_r \times 1}$ and linearly projected to the $\mathbb{R}^{HW \times 1}$ space as $\boldsymbol{\bar{z}}^{(t)} \triangleq \sigma(\boldsymbol{W}_z \boldsymbol{z}^{(t)} +\boldsymbol{b}_z)$, with $H \times W$ being the number of pixels of the desired image, $\boldsymbol{W}_z$ and $\boldsymbol{b}_z$ being trainable parameters of conformable dimensions, and $\sigma(\cdot)$ denotes the sigmoid activation.
Then, $\boldsymbol{\bar{z}}^{(t)}$ is seen as an image $\boldsymbol{\bar{Z}}^{(t)} \in \mathbb{R}^{1\times H \times W}$ and the two computer vision modules are applied.

In the beginning, $L_2=5$ typical ResNet blocks of a single channel are applied on $\boldsymbol{\bar{Z}}$ with the intention of capturing information about the location, size, and shape of the objects, resulting in an intermediate image $\boldsymbol{\bar{I}}^{(t)} \in \mathbb{R}^{1\times H \times W}$.
This information is assumed to exhibit high spatial locality due to the spatially aware graph introduced by the GAT layers, therefore, the convolutional layers of ResNet provide an intuitive choice.
Finally, $\boldsymbol{\bar{I}}^{(t)}$ is processed by a UNet of $L_3 = 3$ downsampling-upsampling layers with $16$, $32$, and $64$ channels, respectively.
The purpose of the U-Net is to efficiently encode information about the shapes of the objects in the image during downsampling to a bottleneck vector, by increasing the channel and decreasing the image dimensions.
The converse procedure of the upsampling layers is designed to decode this information toward assigning each pixel of the image to its corresponding object (or background). Recall that this work deals with perfect conducting objects, indicating that the output image is binary. This implies that it suffices to apply a sigmoid activation to obtain the final image estimation $\boldsymbol{\hat{I}}^{(t)} \in\mathbb{R}^{H \times W} \equiv I(\mathcal{\bar{S}}^{(t)})$, before performing SGD on~\eqref{eq:inverse-problem}.
The overall DNN architecture is illustrated in~Fig.~\ref{fig:dnn-arch}. 

\begin{figure}[t]
    \centering
    \fbox{\includegraphics[width=0.8\linewidth, trim=30 30 30 30, clip]{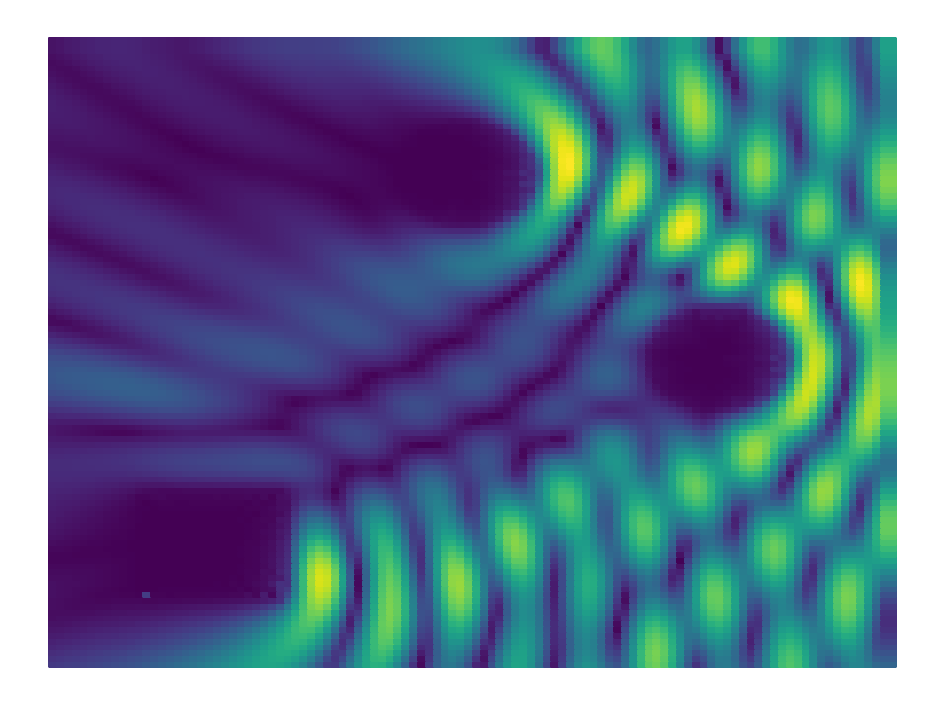}}
    \vspace{-0.1cm}
    \caption{Visualization of the magnitude of the electric field $|E_{\rm r}(\boldsymbol{s})|$ in arbitrary units over the DoI in the presence of three perfect electrical conductors of side and diameter $1.2\lambda$, demonstrating the complex interactions captured by the employed EFIEs. Non-overlapping pulse functions with \(\lambda/20\) spacing in between them were used as bases.}
    \label{fig:EM_model_visualization}
    \vspace{-0.2cm}
\end{figure}

\begin{figure}[t]
    \centering
    \subfloat[MNIST data set]{%
        \includegraphics[width=0.95\linewidth]{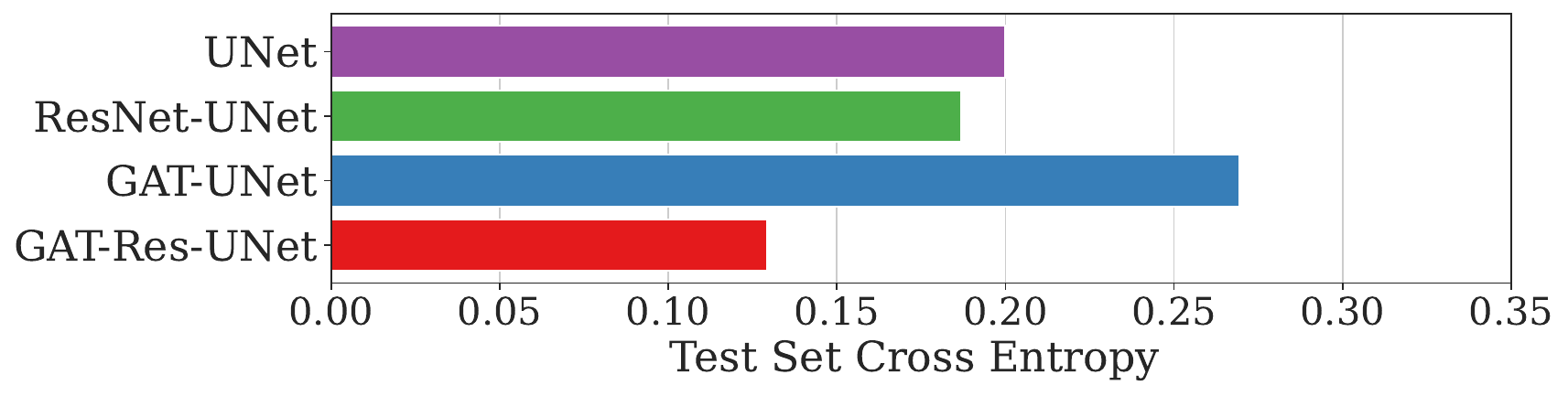}%
        \label{fig:performance-MNIST}
    }
    \vspace{-0.1cm}
    \subfloat[SHAPES data set]{%
        \includegraphics[width=0.95\linewidth]{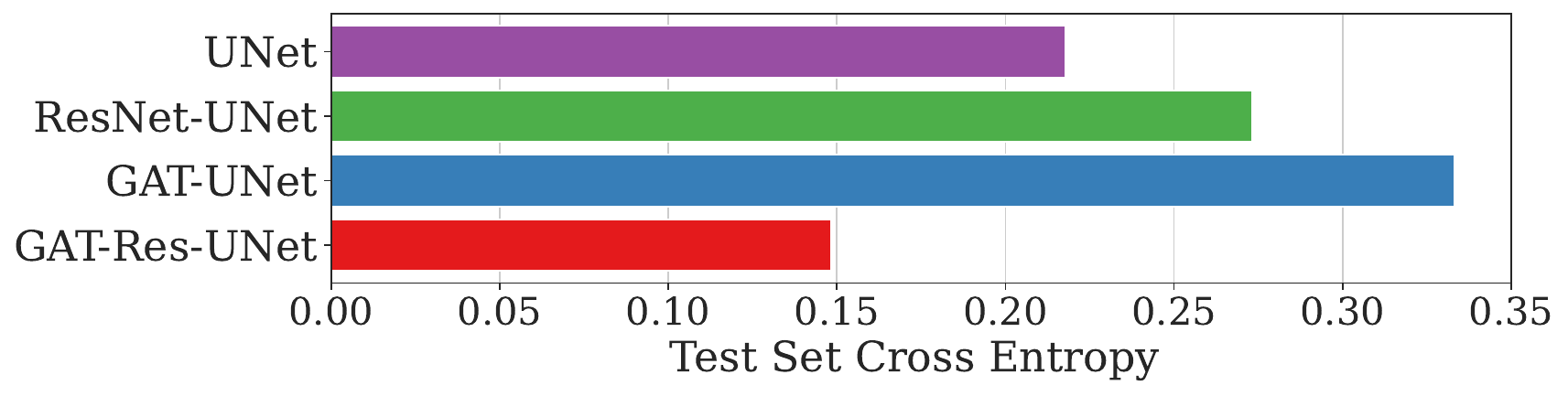}%
        \label{fig:performance-Shapes}
    }
    \caption{Comparison of validation scores for the proposed GAT-Res-UNet and baseline methods on MNIST and SHAPES data sets (lower is better).}
    \label{fig:performance-combined}
\end{figure}

\setlength{\tabcolsep}{4pt}
\begin{table}[t]
    \centering
    \caption{Performance comparison in terms of CE under different SNR levels and $N_r$ values.}
    \label{tab:performance}
    \renewcommand{\arraystretch}{1.2}
    \begin{tabular}{c|cccc|cccc}
        \hline
        \textbf{Data set} & \multicolumn{4}{c|}{\textbf{MNIST}} & \multicolumn{4}{c}{\textbf{SHAPES}} \\
        \hline
        \textbf{SNR (dB)} & $\mathbf{0}$ & $\mathbf{10}$ & $\mathbf{20}$ & $\mathbf{30}$ & $\mathbf{0}$ & $\mathbf{10}$ & $\mathbf{20}$ & $\mathbf{30}$ \\
        \hline

        UNet         & $\ul{0.25}$ & $0.20$ & $0.18$ & $0.19$ & $\ul{0.34}$ & $\ul{0.29}$ & $0.26$ & $0.25$ \\
        ResNet-UNet  & $\ul{0.25}$ & $0.20$ & $0.20$ & $0.20$ & $0.35$ & $0.36$ & $0.37$ & $0.32$ \\
        GAT-Res-UNet & $\ul{0.25}$ & $\ul{0.18}$ & $\ul{0.17}$ & $\ul{0.16}$ & $0.35$ & $0.30$ & $\ul{0.21}$ & $\ul{0.19}$
 \\
        
        \hline
        $\boldsymbol{N_r}$ & $\mathbf{8}$ & $\mathbf{16}$ & $\mathbf{32}$ &$\mathbf{48}$ & $\mathbf{8}$ & $\mathbf{16}$ & $\mathbf{32}$ & $\mathbf{48}$\\
        \hline
        UNet         & $0.27$ & $0.24$ & $0.21$ & $0.19$ & $0.36$ & $\ul{0.34}$ & $0.30$ & $0.31$ \\
        ResNet-UNet  & $0.26$ & $0.25$ & $0.26$ & $0.20$ & $0.37$ & $0.36$ & $0.32$ & $0.28$ \\
        GAT-Res-UNet & $\ul{0.25}$ & $\ul{0.21}$ & $\ul{0.17}$ & $\ul{0.16}$ & $\ul{0.35}$ & $0.36$ & $\ul{0.28}$ & $\ul{0.22}$ \\
        \hline
    \end{tabular}
\end{table}

\begin{figure}[t]
    \centering
    \includegraphics[width=\figwidth]{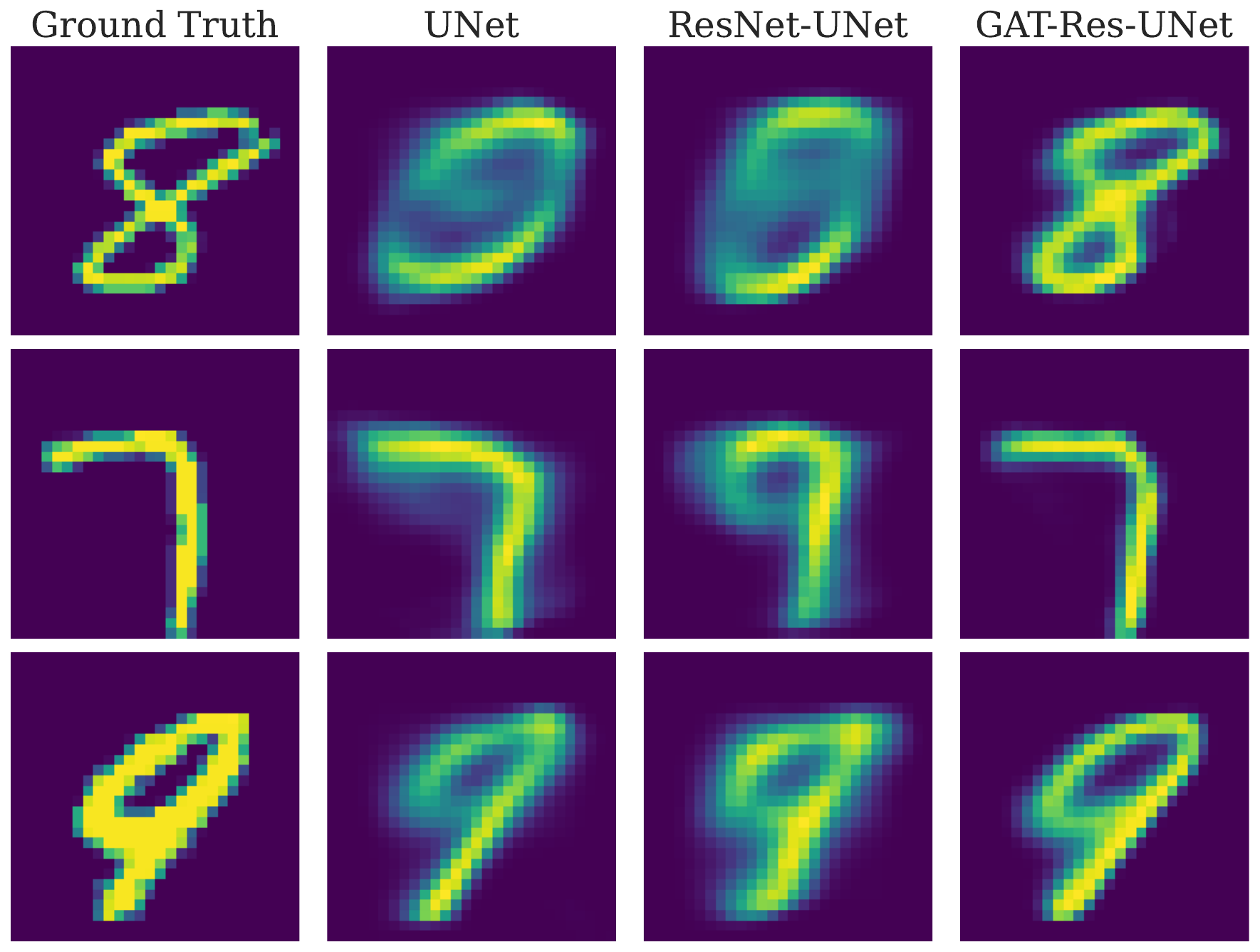}
    \vspace{-0.1cm}
    \caption{Visual comparison of the reconstruction of three MNIST images.}
    \label{fig:reconstructed-MNIST}
    \vspace{-0.2cm}
\end{figure}

\begin{figure}[t]
    \centering
    \includegraphics[width=\figwidth]{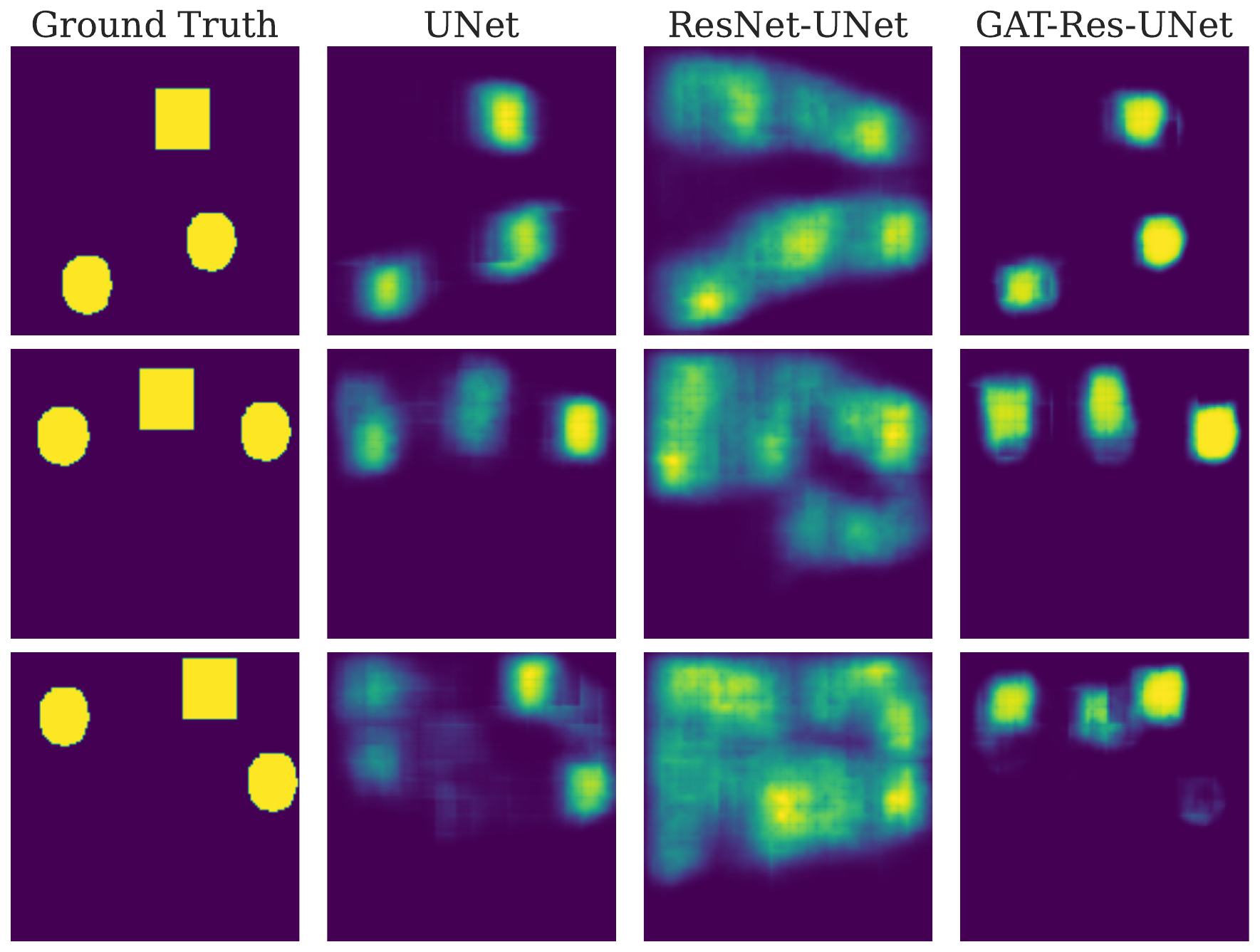}
    \vspace{-0.1cm}
    \caption{Visual comparison of the reconstruction of three SHAPES images.}
    \label{fig:reconstructed-SHAPES}
    \vspace{-0.2cm}
\end{figure}







\begin{figure}[t]
    \centering
    \subfloat[MNIST image under different SNR conditions.]{%
        \includegraphics[width=\figwidth]{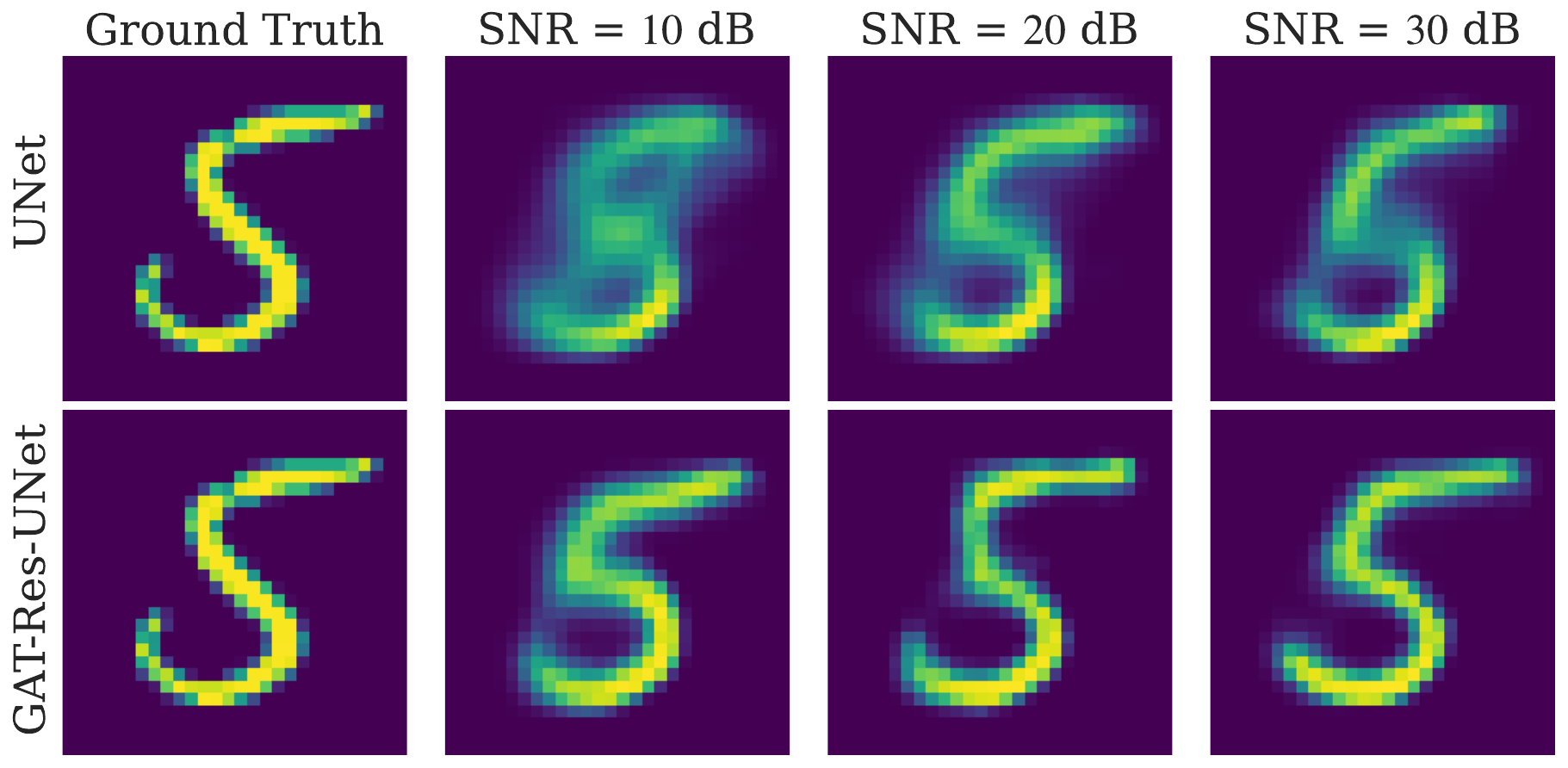}
    \label{fig:mnist-reconstructed-snr}
    }
    \vspace{-0.1cm}
    \subfloat[SHAPES image under different SNR conditions.]{%
        \includegraphics[width=\figwidth]{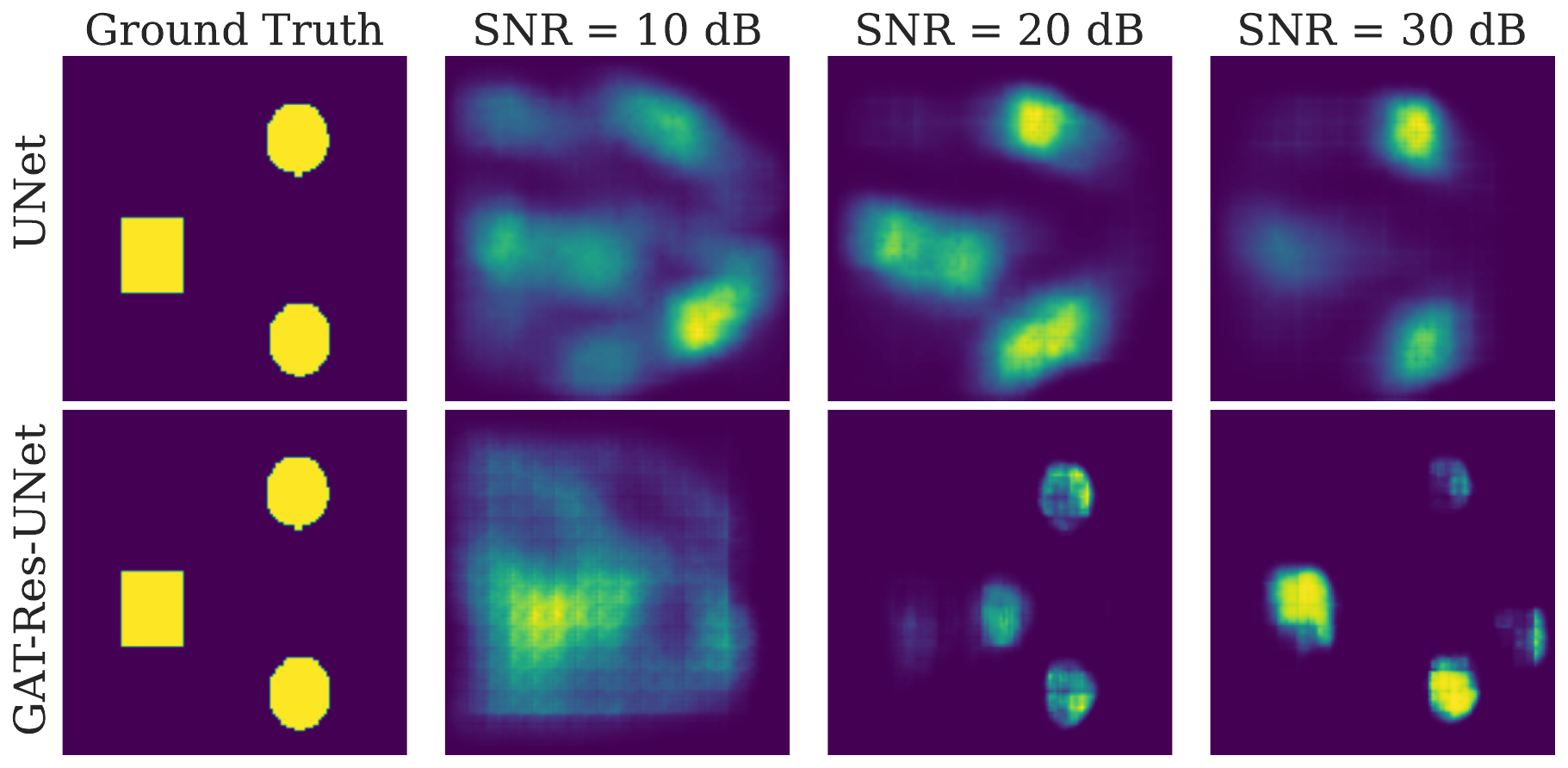}
    \label{fig:shapes-reconstructed-snr}
    }
    \vspace{-0.1cm}
    \subfloat[MNIST image under different numbers of RXs.]{%
        \includegraphics[width=\figwidth]{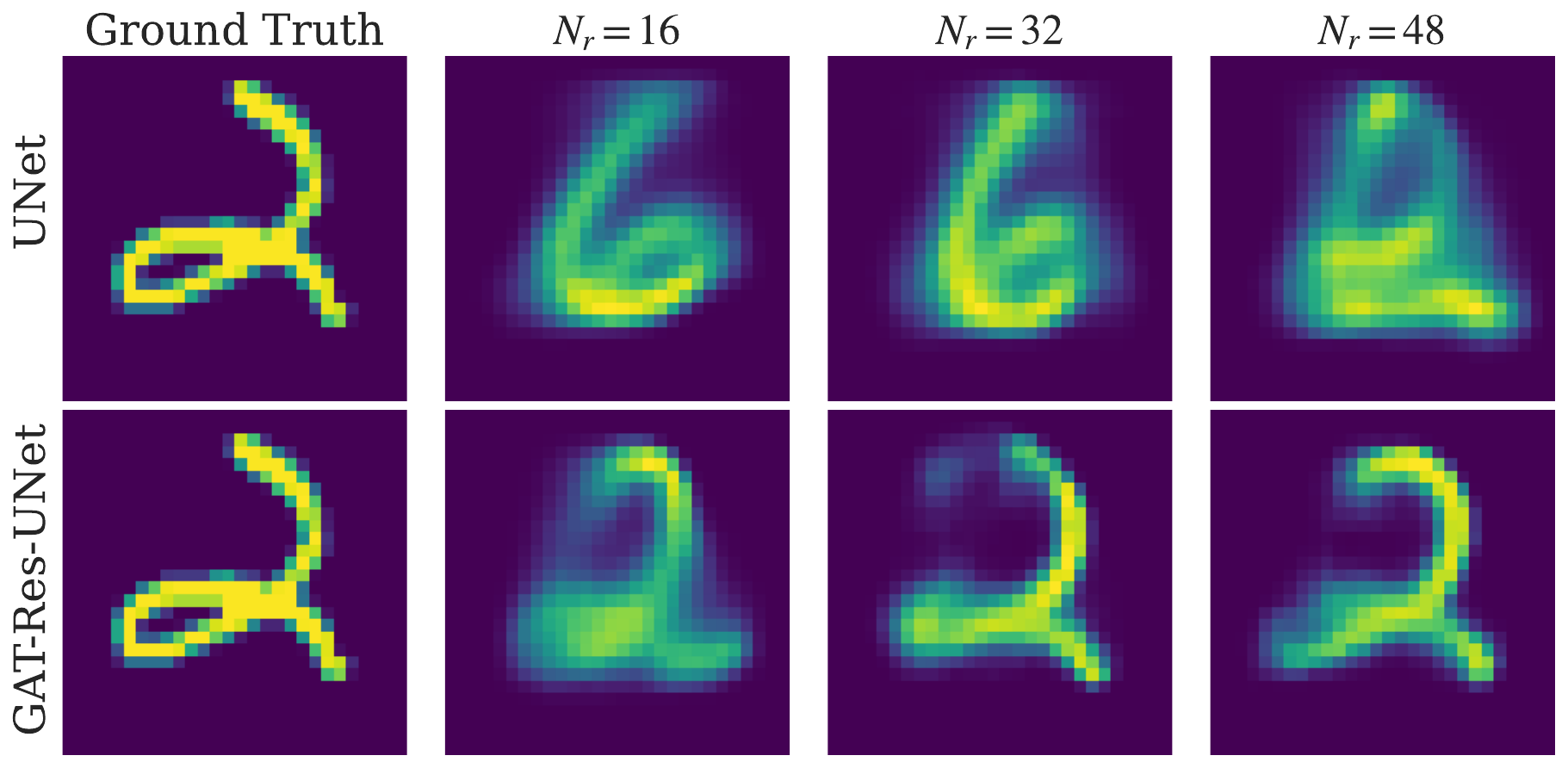}
    \label{fig:mnist-reconstructed-N_r}
    }
    \vspace{-0.1cm}
    \subfloat[SHAPES image under different numbers of RXs.]{%
        \includegraphics[width=\figwidth]{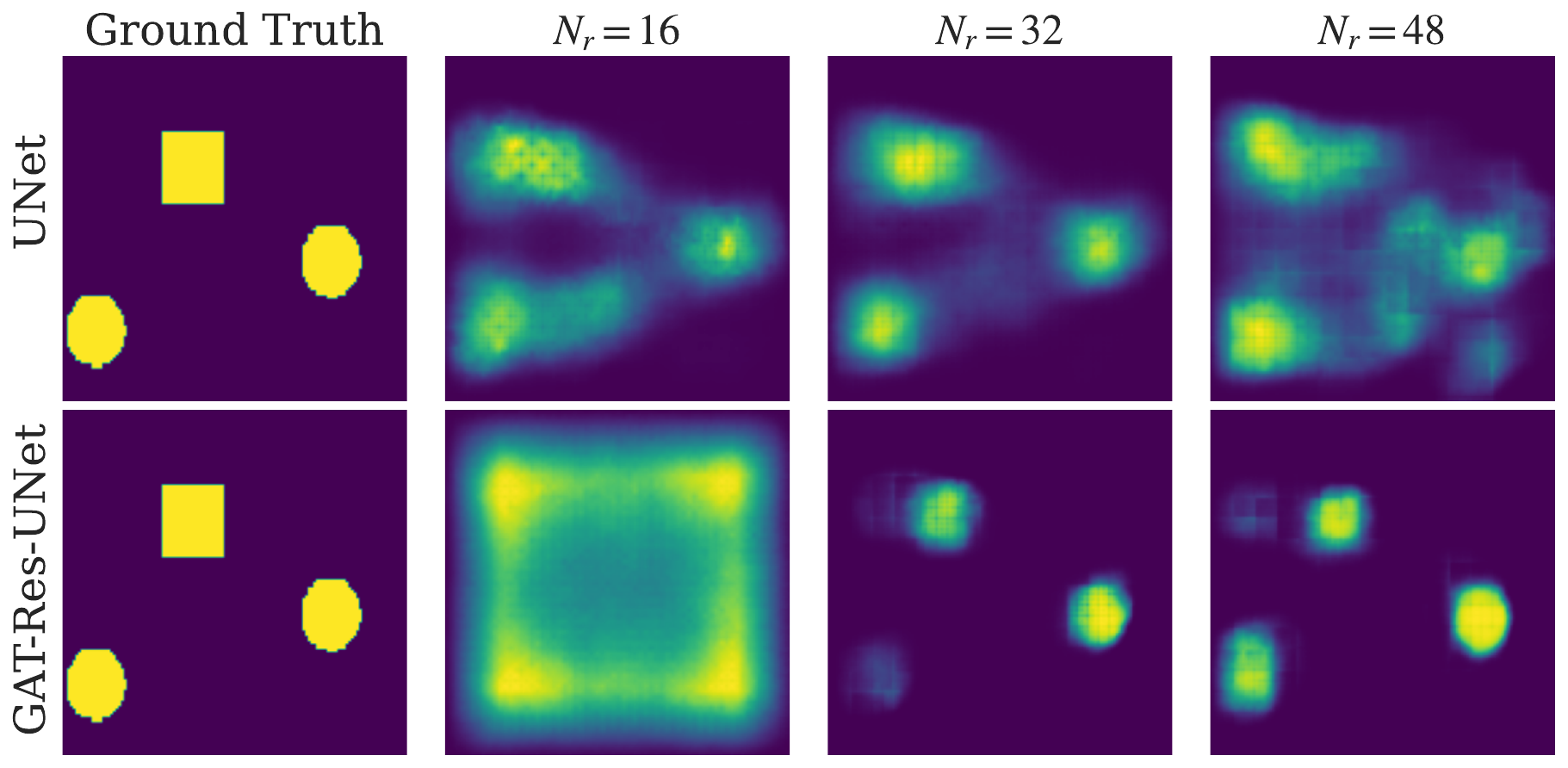}
    \label{fig:shapes-reconstructed-N_r}
    }
    \caption{Reconstruction of sample images under various system parameters.}
    \label{fig:reconstructions-combined}
\end{figure}

\vspace{-0.1cm}
\section{Evaluation}
Using the forward model of Section~\ref{sec:em_model} and $\lambda=0.125~\mathrm{m}$, we have produced two data sets of $10^4$ data points, each with $0.2$ test split, that were designed to test different system characteristics.
First, we sampled MNIST digits~\cite{MNIST} to investigate the ability of the designed DNNs to identify complex shapes.
Each $28\times28$ image was conceptually centered on the DoI origin so that its total height and width are $3\lambda$ each, and its pixel occupies an area of $0.1\lambda \times 0.1\lambda$.
Using the spatial resolution $\rho_0=\lambda/20$, we sampled points within the area occupied by the pixels of the digits, to obtain each $\mathcal{\bar{S}}^{(t)}$ surface, which contains $282$ discrete points on average.
$\mathcal{\bar{S}}^{(t)}$ was then used in~\eqref{eq:EFIEs} to obtain $\boldsymbol{e}^{(t)}$ while setting $N = |\mathcal{\bar{S}}^{(t)}|$ ($|\cdot|$ denotes set cardinality). 
Next, we created a synthetic data set, referred to as ``SHAPES,'' which was designed to assess the detection of multiple disjoint objects with larger image resolution.
To that end, a square and two discs of width and diameters uniformly sampled in $[\lambda, 1.2\lambda]$ were randomly positioned within the DoI disk to create each $\mathcal{\bar{S}}^{(t)}$.
The discretization results in $651$ points on average and the target images had dimensions of $128 \times 128$ pixels.
Figure~\ref{fig:EM_model_visualization} provides a visual inspection of the our EFIE method in the generation of the data set.

We have compared the proposed ``GAT-Res-UNet'' architecture against different architectures of the same three components:
UNet, ResNet-UNet, and GAT-UNet.
The UNet baseline closely follows the benchmark approach of~\cite{UNet_for_Imaging}, while the rest of the approaches constitute modifications of it. 
Other combinations of the same layers did not exhibit notable performance.
For the approaches that do not contain GAT as the first layer, only $\boldsymbol{e}^{(t)}$ was used as input, without the system information of $E_{\rm t}(\boldsymbol{r}_i)$ and $\boldsymbol{D}$.
A linear layer was introduced in these cases instead of GAT to convert the input vectors to the image dimensions similar to the $\boldsymbol{z}^{(t)} \to \boldsymbol{\bar{z}}^{(t)}$ projection of Section~\ref{sec:dnn-arch}, so that the convolutional modules can be applied.

For $N_r=64$ RXs and noise-free received signals, the CE error scores for the test sets of all compared methods are given in Figs.~\ref{fig:performance-MNIST} and \ref{fig:performance-Shapes}.
Clearly, the proposed GAT-Res-UNet approach shows lower reconstruction errors, with the performance difference with respect to the baseline approaches being increased for SHAPES images of higher resolution.
To investigate scenarios where the received signals exhibiting Additive White Gaussian Noise (AWGN), as well as in setups with varying number of RXs, the models were re-trained for each particular SNR and $N_r$ case, with the results being shown in Table~\ref{tab:performance}. As observed, the proposed method outperforms all benchmarks in almost all cases. It is shown that their differences become less pronounced when their resources deteriorate (fewer RXs and lower SNR), where no method produces effective results.
The SHAPES data set presents the more challenging case, due to the higher resolution.

From a qualitative perspective, reconstructed images from the test sets of the two data sets are depicted in Figs.~\ref{fig:reconstructed-MNIST} and \ref{fig:reconstructed-SHAPES}, where it can be seen that GAT-Res-UNet produces sharper images, especially for the case of MNIST.
For the challenging case of SHAPES images, GAT-Res-UNet is mostly able to identify the positions and sizes of the shapes with moderate accuracy, but it does not succeed in producing precise shapes.
Comparing with UNet for different resource budgets in Fig~\ref{fig:reconstructions-combined}, a clear advantage of the proposed method can be inferred in all tests apart from low resources cases, where all methods are unsuccessful.
While there is notable performance degradation as the system capabilities become limited, the proposed method is demonstrated to exhibit higher robustness.

\vspace{-0.1cm}
\section{Conclusion}
In this paper, we designed a DNN architecture based on graph attention, residual convolutions, and UNet structures for effective RF imaging with reduced system requirements. An EM modeling approach based on the EFIEs has been used for generation of two data sets of distinct features, and our numerical evaluation showcased notable performance gains of the proposed method compared to modified approaches from the literature, as well as relative resilience in decreased SNR levels and number of RXs.
Such results further motivate us to investigate the incorporation of the presented EM model to ML methods in making DNN designs physics-informed, toward enhancing their performance and generalization capabilities.

\bibliographystyle{IEEEtran}
\vspace{-0.1cm}
\bibliography{references}

\begin{thebibliography}{10}
\providecommand{\url}[1]{#1}
\csname url@samestyle\endcsname
\providecommand{\newblock}{\relax}
\providecommand{\bibinfo}[2]{#2}
\providecommand{\BIBentrySTDinterwordspacing}{\spaceskip=0pt\relax}
\providecommand{\BIBentryALTinterwordstretchfactor}{4}
\providecommand{\BIBentryALTinterwordspacing}{\spaceskip=\fontdimen2\font plus
\BIBentryALTinterwordstretchfactor\fontdimen3\font minus \fontdimen4\font\relax}
\providecommand{\BIBforeignlanguage}[2]{{%
\expandafter\ifx\csname l@#1\endcsname\relax
\typeout{** WARNING: IEEEtran.bst: No hyphenation pattern has been}%
\typeout{** loaded for the language `#1'. Using the pattern for}%
\typeout{** the default language instead.}%
\else
\language=\csname l@#1\endcsname
\fi
#2}}
\providecommand{\BIBdecl}{\relax}
\BIBdecl

\bibitem{Microwave_Breast_Imaging}
D.~O’Loughlin, M.~O’Halloran, B.~M. Moloney, M.~Glavin, E.~Jones, and M.~A. Elahi, ``Microwave breast imaging: {C}linical advances and remaining challenges,'' \emph{IEEE Trans. Biomed. Eng.}, vol.~65, no.~11, pp. 2580--2590, 2018.

\bibitem{RMA}
D.~Sheen, D.~McMakin, and T.~Hall, ``Three-dimensional millimeter-wave imaging for concealed weapon detection,'' \emph{IEEE Trans. Microw. Theory Tech.}, vol.~49, no.~9, pp. 1581--1592, 2001.

\bibitem{MAIERHOFER2010xv}
C.~Maierhofer \emph{et~al.}, \emph{Non-Destructive Evaluation of Reinforced Concrete Structures}.\hskip 1em plus 0.5em minus 0.4em\relax Woodhead Publishing, 2010.

\bibitem{GPR_book}
H.~M. Jol, \emph{Ground Penetrating Radar Theory and Applications}.\hskip 1em plus 0.5em minus 0.4em\relax Amsterdam, Neverthlands: Elsevier, 2009.

\bibitem{6G-DISAC-magazine}
E.~Calvanese~Strinati \emph{et~al.}, ``Towards distributed and intelligent integrated sensing and communications for {6G} networks,'' \emph{IEEE Wireless Commun.}, vol.~32, no.~1, pp. 60--67, 2025.

\bibitem{MVA24_Metaverse}
A.~Masaracchia \emph{et~al.}, ``Towards the metaverse realization in {6G}: {O}rchestration of {RIS}-enabled smart wireless environments via digital twins,'' \emph{IEEE Internet Things Mag.}, vol.~7, no.~2, pp. 22--28, 2024.

\bibitem{DT_for_Wireless}
L.~Nisiotis \emph{et~al.}, ``Exploring gaming technologies, digital twins, and {VR} to visualise wireless propagation simulations,'' in \emph{Proc. IEEE Annu. Comput., Software, Appl. Conf.}, Osaka, Japan, Jul. 2024.

\bibitem{QUALCOMM_channel_model}
K.~Haneda \emph{et~al.}, ``{5G} {3GPP-Like} channel models for outdoor urban microcellular and macrocellular environments,'' in \emph{Proc. IEEE Veh. Technol. Conf.}, Nanjing, China, May 2016.

\bibitem{BPA_vs_RMA}
G.~Wang \emph{et~al.}, ``Comparison between back projection algorithm and range migration algorithm in terahertz imaging,'' \emph{IEEE Access}, vol.~8, pp. 18\,772--18\,777, 2020.

\bibitem{RMA_MIMO}
Y.~Yang \emph{et~al.}, ``Automotive {SAR} imaging by range migration algorithm,'' in \emph{Proc. Int. Conf. Microw. Millimeter Wave Technol.}, Harbin, China, Aug. 2022.

\bibitem{Yonina_Physics_ML_Imaging}
R.~Guo \emph{et~al.}, ``Physics-embedded machine learning for electromagnetic data imaging: {E}xamining three types of data-driven imaging methods,'' \emph{IEEE Signal Process. Mag.}, vol.~40, no.~2, pp. 18--31, 2023.

\bibitem{Yonina_Deep_Tomographic_Imaging}
G.~Wang \emph{et~al.}, ``Deep tomographic image reconstruction: {Y}esterday, today, and tomorrow—editorial for the 2nd special issue ``{Machine} {Learning} for {Image} {Reconstruction}'','' \emph{IEEE Trans. Med. Imag.}, vol.~40, no.~11, pp. 2956--2964, 2021.

\bibitem{UNet_for_Imaging}
Z.~Wei and X.~Chen, ``Deep-learning schemes for full-wave nonlinear inverse scattering problems,'' \emph{IEEE Trans. Geosci. Remote Sens.}, vol.~57, no.~4, pp. 1849--1860, 2019.

\bibitem{U_Net}
O.~Ronneberger, P.~Fischer, and T.~Brox, ``{U-Net}: {C}onvolutional networks for biomedical image segmentation,'' in \emph{Proc. Med. Image Comput. Comput. Assist. Interv.}, Munich, Germany, 2015.

\bibitem{DL_Imaging_Two_Step}
H.~M. Yao \emph{et~al.}, ``Two-step enhanced deep learning approach for electromagnetic inverse scattering problems,'' \emph{IEEE Antennas Wireless Propag. Lett.}, vol.~18, no.~11, pp. 2254--2258, 2019.

\bibitem{ResNet}
K.~He \emph{et~al.}, ``Deep residual learning for image recognition,'' in \emph{Proc. IEEE Conf. Comput. Vis. Pattern Recognit.}, Las Vegas, NV, USA, 2016.

\bibitem{DL_for_Inverse_Scattering_review}
X.~Chen, Z.~Wei, M.~Li, and P.~Rocca, ``A review of deep learning approaches for inverse scattering problems (invited review),'' \emph{Prog. In Electromagn. Res.}, vol. 167, pp. 67--81, 2020.

\bibitem{Balanis}
C.~A. Balanis, \emph{Advanced Engineering Electromagnetics 2nd Edition}.\hskip 1em plus 0.5em minus 0.4em\relax Hoboken, NJ, USA: John Wiley \& Sons, Ltd, 2011.

\bibitem{GAT}
P.~Veličković \emph{et~al.}, ``Graph attention networks,'' in \emph{Proc. Int. Conf. Learn. Representations}, Vancouver, Canada, May 2018.

\bibitem{pyg_gatconv}
M.~Fey and J.~E. Lenssen, ``{GATConv} --- {PyTorch} {G}eometric 2.4.0 documentation,'' Online, 2024, available: \url{https://pytorch-geometric.readthedocs.io/en/latest/modules/nn.html}.

\bibitem{MNIST}
L.~Deng, ``The {MNIST} database of handwritten digit images for machine learning research,'' \emph{IEEE Signal Process. Mag.}, vol.~29, no.~6, pp. 141--142, 2012.

\end{thebibliography}

\end{document}